\documentclass[letterpaper, 10 pt, conference]{ieeeconf}  

\usepackage[Symbolsmallscale]{upgreek}
\usepackage{xargs}
\usepackage{float}
\usepackage{wrapfig}
\usepackage[normalem]{ulem}
\usepackage{amsmath,amsfonts}
\usepackage{algorithm}
\usepackage{array}
\usepackage{textcomp}
\usepackage{stfloats}
\usepackage{url}
\usepackage{verbatim}
\usepackage{graphicx}
\usepackage{cite}
\usepackage{xcolor}

\usepackage{graphicx}
\usepackage{caption}
\usepackage{subcaption}
\usepackage{svg}
\usepackage{algorithm, setspace}
\usepackage[noend]{algpseudocode}
\usepackage[%
    colorlinks=true,
    pdfborder={0 0 0},
    linkcolor=red
]{hyperref}

\algnewcommand\algorithmicforeach{\textbf{for each}}
\algdef{S}[FOR]{ForEach}[1]{\algorithmicforeach\ #1\ \algorithmicdo}

\usepackage{graphics} 
\usepackage{epsfig} 
\usepackage{mathptmx} 
\usepackage{times} 
\usepackage{amsmath} 
\usepackage{amssymb}  
\usepackage{textcomp, gensymb}
\usepackage{url}
\usepackage{graphicx}
\usepackage{color}
\usepackage{xcolor}
\usepackage{soul}
\usepackage{mathtools}
\usepackage{algorithm, setspace}
\usepackage[noend]{algpseudocode}
\usepackage{ccicons}
\usepackage{comment}
\usepackage{bbm}
\usepackage{multirow}
\usepackage[compact]{titlesec}

\usepackage[font={small}]{caption}
\renewcommand{\baselinestretch}{0.97}

\newcommand{\state}{\mathbf{x}}

\newcommand{\ctrl}{u}

\newcommand{\rhplanner}{RHPlanner}
\newcommand{\network}{SPARQ }

\newtheorem{remark}{Remark}
\IEEEoverridecommandlockouts                              

\title{\LARGE \bf
System-Level Safety Monitoring and Recovery for Perception Failures in Autonomous Vehicles
}

\author{Kaustav Chakraborty$^{*1}$, Zeyuan Feng$^{*1}$, Sushant Veer$^{2}$, Apoorva Sharma$^{2}$,\\ Boris Ivanovic$^{2}$, Marco Pavone$^{2}$, Somil Bansal$^{1}$  
\thanks{*equal contribution. Corresponding author: kaustavc@usc.edu.}
\thanks{$^{1}$Authors are with Department of Electrical and Computer Engineering,  University of Southern California, Los Angeles, USA.}
\thanks{$^{2}$Authors are with NVIDIA Research
        }%
\thanks{ This work is supported in part by the DARPA Assured NeuroSymbolic Learning and Reasoning (ANSR) program and by the NSF CAREER program (2240163).}
}

\begin{document}

\maketitle
\thispagestyle{empty}
\pagestyle{empty}

\begin{abstract}
The safety-critical nature of autonomous vehicle (AV) operation necessitates development of task-relevant algorithms that can reason about safety at the system level and not just at the component level. To reason about the impact of a perception failure on the entire system performance, such task-relevant algorithms must contend with various challenges: complexity of AV stacks, high uncertainty in the operating environments, and the need for real-time performance. To overcome these challenges, in this work, we introduce a Q-network called SPARQ (abbreviation for \underline{S}afety evaluation for \underline{P}erception \underline{A}nd \underline{R}ecovery \underline{Q}-network) that evaluates the safety of a plan generated by a planning algorithm, accounting for perception failures that the planning process may have overlooked. This Q-network can be queried during system runtime to assess whether a proposed plan is safe for execution or poses potential safety risks. If a violation is detected, the network can then recommend a corrective plan while accounting for the perceptual failure. We validate our algorithm using the NuPlan-Vegas dataset, demonstrating its ability to handle cases where a perception failure compromises a proposed plan while the corrective plan remains safe. We observe an overall accuracy and recall of \textbf{90\%} while sustaining a frequency of \textbf{42Hz} on the unseen testing dataset. We compare our performance to a popular reachability-based baseline and analyze some interesting properties of our approach in improving the safety properties of an AV pipeline. \\
Website: {\tt\footnotesize \url{vatsuak.github.io/sparq}}
\end{abstract}
\section{Introduction}
\label{sec:intro}
\begin{figure*}[t]
\vspace{0.3em}
  \centering
  \includegraphics[width=0.89\textwidth]{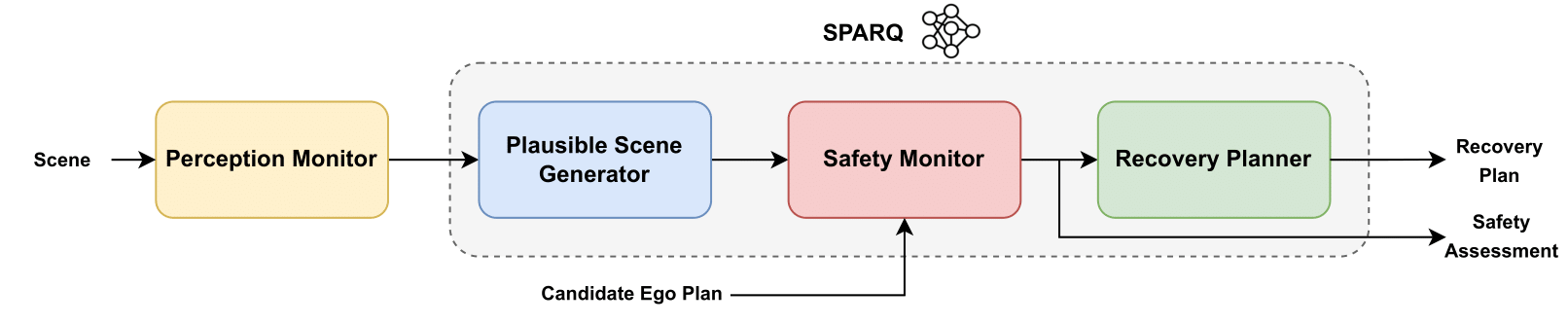}
  \vspace{-2mm}
  \caption{Block diagram for task-relevant perception failure detection and recovery monitor. SPARQ learns a Q-network-based neural approximator for the last three blocks in this pipeline enabling it to operate at run-time at a frequency of $\sim$42 Hz.}
  \label{fig:block-diagram}
  \vspace{-3.5mm}
\end{figure*}

Safety of autonomous vehicles (AVs) heavily relies on the correctness of perception systems. This has spurred a plethora of work in developing monitors aimed at detecting perception failures \cite{antonante22monitoring,Balakrishnan21-percemon,miller2022what,rahman2022fsnet}. However, triggering a fail-safe maneuver for every detected perception error is impractical and detrimental to the AV's navigation goals. Not all perception failures are equal; some bear no impact on the AV's safety (e.g., missed parked vehicle far off from the AV's motion plan), while others can be catastrophic (e.g., missed pedestrian along the AV's motion plan); the latter are referred to as task-relevant (or system-level) perception failures \cite{antonante2023task}. Identifying task-relevant perception failures and reacting to them poses significant challenges: First, the AV operates in highly uncertain environments, making it difficult to assess the system-level impact of perception failures. Second, these monitors must assess the system-level impact of complex AV stacks at run-time, ideally at a much higher frequency than the AV stack. Finally, even if the perception failure is detected, there might not be enough time to replan a safe motion accounting for the perception errors. In this work, we address these challenges by developing a Q-network-based monitor called SPARQ (abbreviation for \underline{S}afety evaluation for \underline{P}erception \underline{A}nd \underline{R}ecovery \underline{Q}-network) that, in real-time, detects task-relevant perception failures and generates corrective recovery plans.
\begin{figure}[t!]
    \centering
\includegraphics[width=\columnwidth]
         {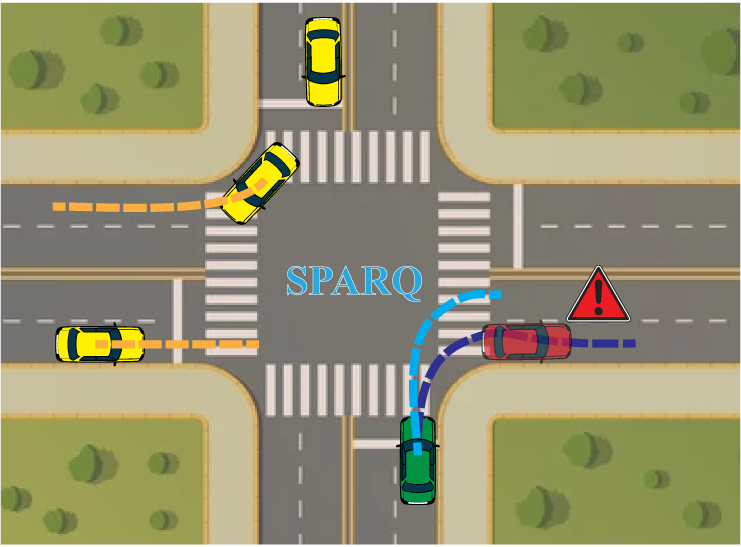}
        \caption{\small{
        We propose SPARQ, a run-time system-level safety monitor to detect and mitigate task-relevant perception failures.
        Here, the perception suite failed to detect the red car. This resulted in AV (the green car) planning an unsafe trajectory (blue line). SPARQ flags the planned trajectory as unsafe and suggests a corrective safe plan (cyan line) that accounts for the perception failure.}}
         \vspace{-1.5em}
         \label{fig:cover_pic}
\end{figure}
We envision a task-relevant perception monitoring and recovery pipeline that has four main blocks: a component-level \emph{perception monitor} that identifies the failure mode that is occurring, a \emph{plausible scene generator} which produces plausible scenes accounting for the perception failure, a \emph{safety monitor} which quantifies how risky it is to execute a given motion plan in the plausible scenes while accounting for the uncertain future predictions of other agents, and finally, a \emph{recovery planner} which generates a recovery maneuver in the event of a task-relevant perception failure (see Fig.~\ref{fig:block-diagram} for the block diagram). Unfortunately, perception monitors have very stringent compute requirements and need to run at frequencies much higher than that of the AV stack, which makes it infeasible to run such an intricate monitoring pipeline online -- indeed, the slow speed of the modular monitoring pipeline was something we observed in our previous work \cite{antonante2023task} as well. SPARQ resolves this problem by learning a neural approximation of the last three blocks, providing run-time capabilities for our task-relevant perception monitoring and recovery pipeline. SPARQ leverages a lightweight transformer-based encoder-decoder architecture and is trained via supervised learning. To generate the training data, we construct the plausible scene generator, safety monitor, and recovery planner and run them offline. In our experiments, we observe that SPARQ is able to provide low false-positive and false-negative detection rates of \textbf{8.1\%} and \textbf{8.5\%}, respectively, while being able to execute at a frequency of \textbf{42} Hz.

\textbf{Statement of Contributions.} Our main contributions are: (i) We develop SPARQ, a Q-function-based run-time capable system-level safety monitor that detects task-relevant perception failures and \emph{simultaneously} plans recovery motions for them. (ii) In the absence of perception failures, we show that SPARQ acts as a general-purpose safety monitor for motion plans, i.e., it detects if a plan is unsafe to execute. (iii) Finally, we demonstrate the ability of SPARQ to assess the system-level impact of perception failures and generate recovery plans on nuPlan \cite{nuplan}, a real-world driving dataset.

\section{Related Works}
\label{sec:related-work}
\textbf{Safety for AVs.} Due to the safety-critical nature of AV operation, a lot of research has focused on ensuring the system-level safety of AVs. These approaches can broadly be classified under two overarching notions: safety under worst-case and probabilistic notions of safety. Approaches within the former category often take the form of control invariant sets within the state space; some notable examples include Hamilton Jacobi (HJ)-reachability \cite{bansal2017hamilton}, control-barrier functions \cite{ames2019control}, collision-free forward reachable sets \cite{manzinger2020using}, and control contraction metrics \cite{singh2017robust}. Approaches within the latter category use probabilistic notions of safety such as estimating collision probability using trajectory prediction \cite{farid22taskrelevant}, conditional value at risk (CVaR) \cite{nyberg2021risk}, and conformal prediction \cite{lindemann2023safe}. Various metrics have also been used for assessing safety, many of which can be found in \cite{westhofen21criticalityMetrics}. Unlike this paper, the above methods assess system-level safety under the assumption of perfect perception.

\textbf{Safety Under Perception Failures.} There has been a significant amount of work on monitoring perception systems, such as \cite{antonante22monitoring,Balakrishnan21-percemon}, and also on identifying specific perception failures such as object detection \cite{miller2022what} and semantic segmentation \cite{rahman2022fsnet}, among many others. However, methods for assessing the system-level impact of perception failures is still a growing topic. Studies like \cite{bansal2021risk}, introduced risk-ranked recall for object detection, and \cite{bernhard2022risk}, created risk envelopes based on perception uncertainty, but both overlook predicting other agents' future actions. Some recent approaches have used HJ reachability zones to account for the potential future actions of other agents to detect perception failures \cite{topan2022interaction,chakraborty2023discovering}. However, HJ reachability-based methods are overly conservative due to worst-case assumptions on other agent behavior. Antonante et al. \cite{antonante22monitoring} propose replacing this worst-case assumption by leveraging learned trajectory prediction models to only consider statistically likely behaviors for other agents, but this method is too computationally heavy for online operation. 
SPARQ takes a similar probabilistic approach when considering agent behavior, but only explicitly generates the future trajectories of agents at train time, distilling the final safety assessment of a candidate plan into a single lightweight neural network.

\textbf{Recovery Planning.} The controls literature is rich with the class of algorithms known as safety filters which keep an autonomous system within a safe set by generating corrective actions whenever the system's state approaches the ``edge" of the safe set; see \cite{wabersich2023data} for an overview on safety filters. These approaches, though powerful and accompanied by safety guarantees, are often overly conservative and struggle to incorporate the rich information and uncertainty rife in autonomous driving. Some approaches that address fallback planning for perception-driven uncertainties include shield model-predictive path integral (MPPI) \cite{yin2023shield} and hand-crafted fallback policies \cite{gupta2024detecting}. However, both these approaches are applied to single-agent settings unlike autonomous driving. Other recent approaches include game-theoretic trajectory repair \cite{wang2024interaction}, satisfiability modulo theory (SMT)-based repair \cite{lin2022rule}, and leveraging Large Language Models (LLMs) \cite{lin2024drplanner}. Notably, none of these approaches operate under perception failures. Unlike the prior literature, SPARQ generates recovery plans for autonomous driving in dense multi-agent scenes while accounting for perceptual errors.

\section{Problem Setup}
\label{sec:problem}
 We consider an autonomous vehicle (referred to as the ``ego") operating as a dynamic system on a roadway, equipped with sensing and planning modules. We assume access to the map data showing pedestrian crossing, road lanes etc, which we term as road graph elements.  In addition to static elements, the ego shares its environment with other vehicles, whose states are accurately estimated up to the current moment. The task of the ego is to utilize its autonomous capabilities to plan and execute dynamically feasible trajectories to achieve the task of navigation while ensuring safety concerning its surroundings. We call this as safe operation of the AV.

 However, the ego's sensors are susceptible to performing sub-optimally and producing erroneous data, such as missed agents, leading to inaccurate predictions and potentially flawed planning. To mitigate this issue of \textit{perception failure}, a real-time perception monitor is used to detect discrepancies between the sensed data and the actual environment. Additionally, we assume access to a reliable ``offline'' algorithm that evaluates the safety of any trajectory proposed by the base planner within the AV's navigation stack.
 
The challenge arises when the perception monitor detects an error in the sensor feedback during runtime. In such cases, it becomes difficult to determine whether the currently planned trajectory (the ``candidate plan") can still ensure safe operation. 
Thus, it is essential to possess an algorithm that can evaluate the candidate plan’s safety under the presence of perception errors \emph{online}, and if the plan is deemed unsafe, the algorithm should correct the plan in real-time based on feedback from the perception monitor.

Hence, our key objectives in this work are:\\
\textbf{Online Safety Evaluation of Candidate plan Under Perception Failure:} Develop an \textit{online} algorithm that integrates real-time feedback from the perception monitor to evaluate the safety score of any candidate plan.

\textbf{Plan Repair}: Design an algorithm to adjust and repair any candidate plan to ensure safety, taking into account both the faulty sensor observations and the corrections provided by the perception monitor.

\section{Background: RH Planner}
\label{sec:background}
We will now go over a quick description of the rule hierarchy (RH) planner \cite{veer2023rulehierarchies}, which we use as our offline recovery planner for data collection.
The RH Planner generates a trajectory tree and then evaluates a reward $R$ for each branch of the tree. The trajectory with the highest reward is chosen for execution. The reward function encompasses various traffic rules in the form of a rule hierarchy \cite{veer2023rulehierarchies} wherein each rule is expressed as a Signal Temporal Logic (STL) formula. Rules in a rule hierarchy are assigned an importance ranking -- the key distinguishing property of rule hierarchies is their ability to relax less important rules in favor of the more important ones in the event that all rules cannot be simultaneously satisfied. Rule hierarchies are a powerful tool to prescribe criteria for over-specified systems where multiple criteria exist that may conflict with one another, e.g., violate speed limit to avoid collision. The rule hierarchy reward $R$ measures how well a trajectory adheres to the rule hierarchy, i.e., trajectories that satisfy more important rules receive a higher reward than those that only satisfy less important rules. In this paper, our rule hierarchy comprises of the following rules in decreasing order of importance: (i) \texttt{collision avoidance}; (ii) \texttt{offroad avoidance}; (iii) \texttt{traffic light compliance}; (iv) \texttt{speed limit compliance}; (v) \texttt{progression}; (vi) \texttt{stay near the route plan center line}; and (vii) \texttt{stay aligned with the route plan center line}. Finally, we note that we use a lane-keep predictor to forecast the future of non-ego vehicles, while we use a constant-velocity predictor for pedestrians.

\section{System-Level Safety Monitoring and Recovery}
\label{sec:approach}
\begin{figure}[ht]
    \centering
    \vspace{0.5em}
         \includegraphics[width=\columnwidth]
         {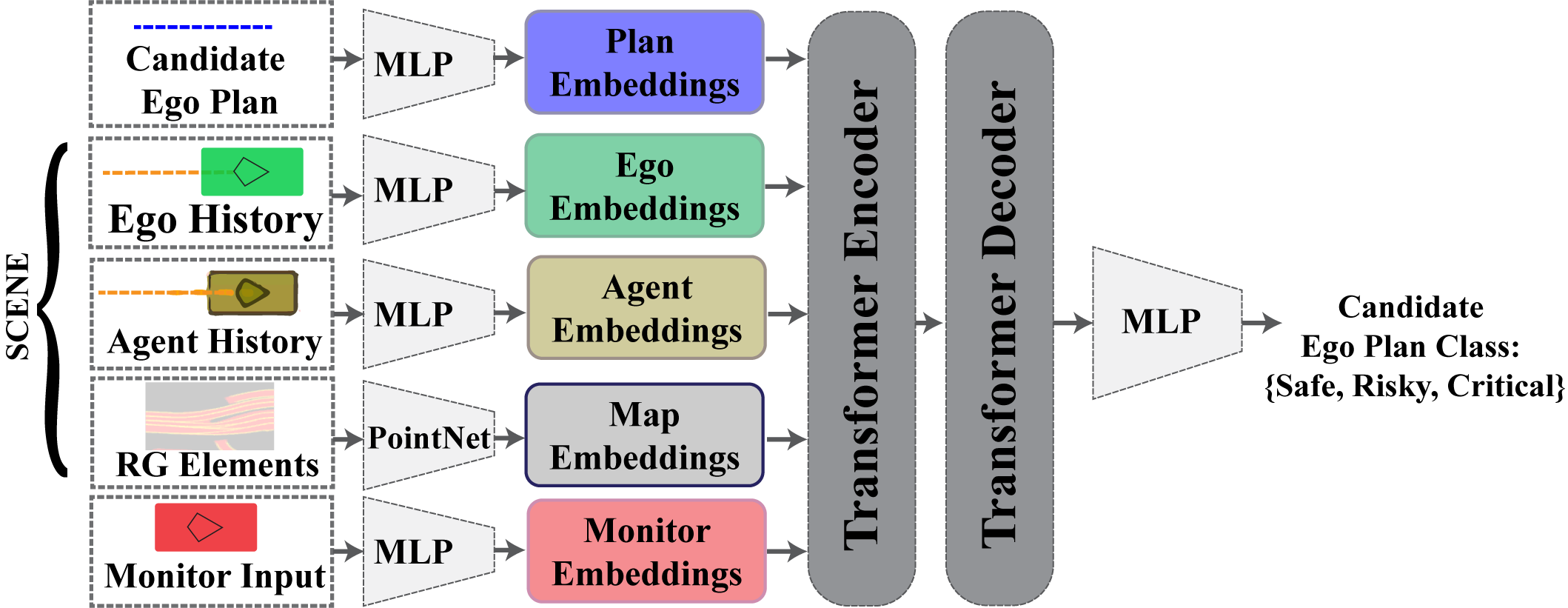}
        \caption{\small{The SPARQ Network Structure}}
 \vspace{-0.75em}
 \label{fig:sparq_structure}
\end{figure}
SPARQ takes as input the (i) scene information, which includes the state history of the ego and surrounding agents and the road graph (RG) elements; (ii) output of a component-level perception monitor which indicates the nature of a potential perception error if one exists, such as the estimated position and velocity of a missed agent; and (iii) a candidate ego plan. The network then outputs a safety score for the plan and recovery plans if the original plan is unsafe. A detailed architecture of SPARQ is provided in Fig.~\textbf{\ref{fig:sparq_structure}}. In the rest of this section we will discuss the various aspects involved in developing SPARQ.

\subsection{Safety Score}
SPARQ predicts the safety score of a given motion plan, which quantifies how risky the plan is to execute. Our approach can work with any base safety score, such as time-to-collision or distance to the nearest agent, but we use the rule-hierarchy reward function presented in Section~\ref{sec:background} for its ability to tie in multiple safety criteria. 
 
However, on its own, these base scores can only assess the risk of a motion plan relative to perceived map elements and obstacles, and fail to account for safety violations due to interactions with scene elements that the AV perception system did not detect at runtime. 
Runtime perception monitors can often provide some information about potential perception failures, for example, roughly localizing a potentially missed obstacle in the scene \cite{antonante22monitoring}. 
In SPARQ, we develop a module which adapts a base safety score into a safety score which can account for information provided by a runtime perception monitor.
The key advantage of SPARQ is its fast inference speed, which enables online evaluation of safety-critical violations under the current plan while accounting for the perception error information from perception monitors.

\begin{remark}
    Note that since the predicted safety score is conditioned on the ego plan, notionally, SPARQ is reminiscent of a Q-function, where the perceived scene and the perception failure information comprise the observation, and the ego plan is the action. Therefore, we refer to it as a Q-network.
\end{remark}

\subsection{Scene Encoding}
As input, \network takes the scene, the candidate plan for the ego, and the monitor input, transforming them into a latent embeddings. The road graph, represented by point clouds of various map elements, is processed using the PointNet \cite{qi2017pointnet++} architecture to extract relevant features, while other inputs are processed via Multi-Layer Perceptrons (MLPs). Both spatial and temporal encodings are added to these features, resulting in a unified embedding for each input type. 

To further refine these embeddings, we employ a transformer-based attention mechanism. Specifically, the model iterates self-attention over the plan embeddings and cross-attention to the embeddings of other elements (e.g., ego, agents, road graph elements and monitor). To optimize computational efficiency, we leverage a Multi-Axial Attention \cite{ho2019axial} mechanism, alternatively performing attention over timesteps and attention across elements, which significantly reduces the complexity relative to attending across all dimensions simultaneously.

The encoded embeddings are then passed through a decoder that uses self-attention to refine the latent representation. Finally, an aggregation MLP projects the latent representation to a single vector representing logits for the quantized values of the safety score.

\subsection{Dataset}
\begin{figure*}[ht]
    \vspace{0.5em}
    \centering
         \includegraphics[width=0.8\textwidth]
         {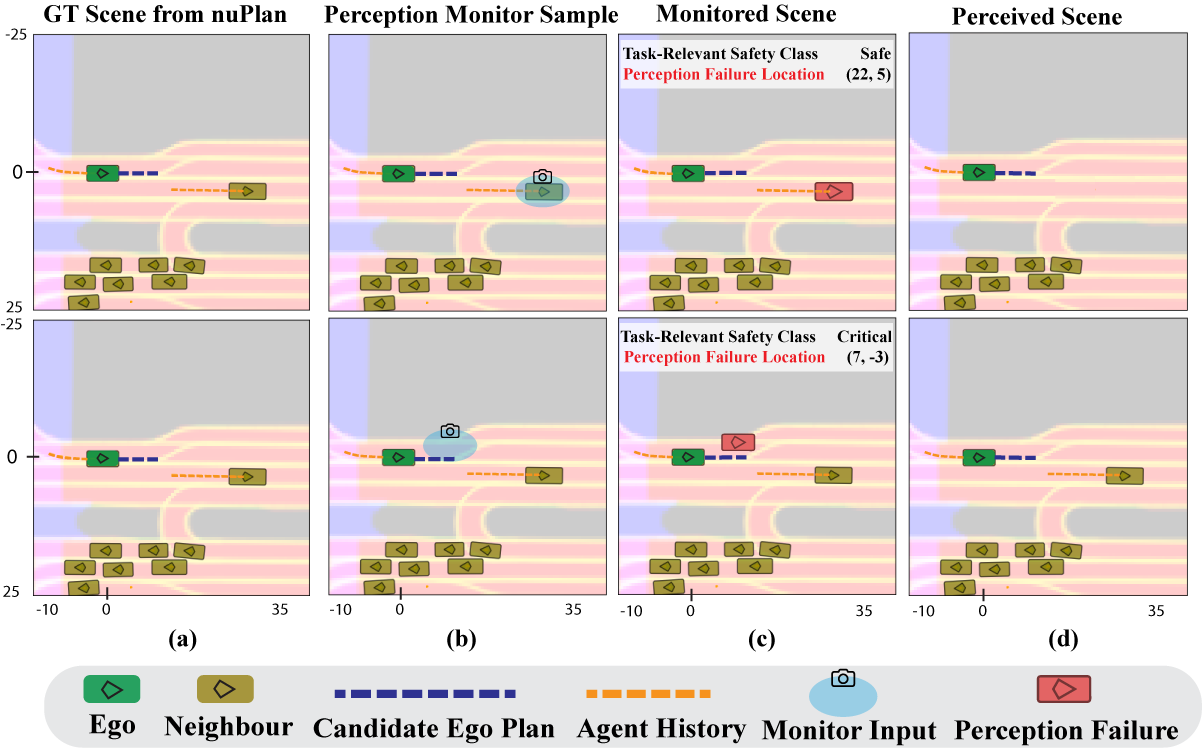}
        \caption{\small{ Data Generation Structure. As our first step we sample a random position as our monitor input. The \textbf{top row} shows the case when an agent exists at the sampled position of the perception monitor (column (b)). We remove the agent from the ground truth (GT) scene to construct our perceived scene (column (d)).  The \textbf{bottom row} shows the case when an agent does not exist. In such cases we use the GT scene as the perceived scene. In either case, our training sample comprises of (perceived scene, ego candidate plan), while the corresponding class label is obtained by evaluating the plan in the monitored scene (column (c)).}}
         \vspace{-0.75em}
         \label{fig:data_collection}
\end{figure*}

SPARQ is trained using supervised learning using existing AV datasets like nuPlan \cite{nuplan}. Each training sample consists of a scene, a planned trajectory, and a monitor input simulating a missed agent. The dataset collection process is depicted in Fig. \ref{fig:data_collection} and outlined in Algorithm \ref{alg:dataset}.

We first sample a scene $s$ (Algorithm:Line \ref{alg:dataset_samplescene}, Fig. \ref{fig:data_collection}a) from the nuPlan dataset.  Next, we simulate a perception failure at a random state $p_{mon}$ (assumed to be normally distributed $N(\mathbf{0,I})$, Algorithm:Line \ref{alg:dataset_samplefailure}, Fig. \ref{fig:data_collection}b). 
This can give rise to two possible scenarios:\textbf{(i)} There are no agents near the failure (Algorithm:Line \ref{alg:dataset_nonearagent}, Fig. \ref{fig:data_collection} \textbf{bottom row}), \textbf{(ii)} There are agents near the perception failure. (Algorithm:Line \ref{alg:dataset_nearagent}, Fig. \ref{fig:data_collection} \textbf{top row})
Depending on the case, we manually edit the scene to simulate the perception failure and obtain the monitored scene (e.g., Fig. \ref{fig:data_collection}c) and the corresponding perceived scene (e.g., Fig. \ref{fig:data_collection}d). We then \texttt{Plan} the planned trajectory tree (Algorithm:Line \ref{alg:dataset_plan_tree}) on the perceived scene and \texttt{Evaluate} the safety reward by using the rule-based scoring system of RHPlanner on the monitored scene  (Algorithm:Line \ref{alg:dataset_rewards}). The rewards are finally categorized into class labels. 

To maintain dataset realism, it is crucial to balance the randomness of the monitor input while ensuring that the edits to the scene remain plausible. To achieve this, we prioritize the sampling of the failure states near existing agents in the scene (\textbf{top row} in Fig. \ref{fig:data_collection}). This approach reduces the need to introduce artificial information about agent behavior or account for interaction dynamics between newly introduced or missing agents, as the monitored agents were already part of the original scene; consequently, the monitored scenes (Fig. \ref{fig:data_collection}c \textbf{top row}) are similar to the raw ground truth scene (Fig. \ref{fig:data_collection}a \textbf{top row}) which are obtained using real-world driving logs. 

\subsection{Training}
Training SPARQ using this dataset in a self-supervised approach takes around 2 days on an NVIDIA RTX 4090 GPU for 25 epochs, with batches of 256 scenarios per iteration. We use the Adam optimizer with a learning rate of \texttt{$2\times10^{-5}$} and a linear decay schedule.

\begin{remark}
    Even though our safety monitor predicts a discrete safety class, the proposed framework can readily be adapted to regress a continuous safety score instead. 
\end{remark}

\begin{algorithm}[h!]
\caption{Dataset Collection for Training Safety Monitor}\label{alg:dataset}
\begin{algorithmic}[1]
\State \textbf{Input:} number of samples $N$, raw dataset $D_{raw}$, e.g., the Nuplan dataset, planner $Plan$, rule-based evaluator $Eval$
\State \textbf{Output:} Monitor dataset $\mathcal{D}$
\State \textbf{Initialize} $\mathcal{D} \leftarrow \emptyset$
\For{\texttt{i = \{1,2...,N\}}}
    \State Sample a scene $s \in \mathcal{D}_{raw}$ \label{alg:dataset_samplescene}
    \State all\_agents $\leftarrow get\_agents(s)$ \Comment{Get all agents}
    \State $p_{mon} \sim N(p_{mon}; \mathbf{0}, \mathbf{I})$ \Comment{Sample failure} \label{alg:dataset_samplefailure}
    \State nearby\_agents $\leftarrow$ $agents\_near(p_{mon})$ 
    \If {all\_agents = $\emptyset$}  \Comment{s has no agents} \label{alg:dataset_noagent}
        \State failure $\leftarrow$ random state in s
        \State $\text{perceived}_{scene}$ $\leftarrow s$
        \State $\text{monitored}_{scene}$ $\leftarrow s + \text{failure}$ \Comment{add failure}
    \ElsIf {nearby\_agents $= \emptyset$} \Comment{No nearby agents} \label{alg:dataset_nonearagent}
        \State agent $\sim$ all\_agents \Comment{random sample}
        \State failure $\leftarrow get\_state(agent)$
        \State $\text{perceived}_{scene}$ $\leftarrow s$
        \State $\text{monitored}_{scene}$ $\leftarrow s + \text{failure}$ \Comment{add failure}
    \ElsIf {nearby agents $\neq \emptyset$} \label{alg:dataset_nearagent}
        \State agent $\sim$ nearby\_agents \Comment{random sample}
        \State failure $\leftarrow get\_state(agent)$
        \State $\text{perceived}_{scene}$ $\leftarrow s - agent$ \Comment{remove agent}
        \State $\text{monitored}_{scene}$ $\leftarrow s$
    \EndIf
    \State plans $\leftarrow$ $Plan$($\text{perceived}_{scene}$) \label{alg:dataset_plan_tree}
    \State label $\leftarrow$ $Evaluate$(plans, $\text{monitored}_{scene}$) \label{alg:dataset_rewards}
    \State $\mathcal{D} \leftarrow \mathcal{D} \cup \left\{ \text{($\text{perceived}_{scene}$, failure, plans, label)} \right\}$
\EndFor
\end{algorithmic}
\end{algorithm}

\subsection{Recovery Planning}

Unlike the modular pipeline presented in Fig.~\ref{fig:block-diagram}, SPARQ can produce safety evaluation as well as a recovery plan simultaneously. Owing to the ability to parallelize computations in a neural network, \network allows us to evaluate a batch of proposed plans including the candidate ego plan under the monitor input in a single forward pass of the network. The candidate plans can either be generated for this purpose online or alternate plan candidates can directly be ported over from the original planner. In this paper, we use the trajectory tree generator from \cite{chen2023interactive} which leverages the differential flatness of vehicular dynamics and GPU parallelization to rapidly produce the tree during run-time. If the candidate ego plan is evaluated as unsafe, we perform plan repair by using one of the proposed plans in the batch that are deemed to be safe by SPARQ, otherwise we return the candidate plan as our safe plan. Since \network incorporates the effects of perception failures, any plan classified as safe by the network is robust to the input perception failures.

\section{Experimental Results and Discussion}
\label{sec:results}
In this section, we demonstrate the following key findings: (i) SPARQ outperforms the reachability-based baseline on all metrics demonstrating better balance of safety and performance; (ii) in lieu of perception errors, SPARQ acts as a general-purpose safety filter; and (iii) SPARQ is run-time capable with an operating frequency of $\sim$42 Hz.

\subsection{Test Datasets and Scenarios} 
We use the nuPlan-Vegas dataset \cite{nuplan} to showcase our proposed approach. Each data sample has been modified to include a perception failure detected by the monitor. Since the base planner is unaware of this failure, the reward of the planned trajectories may degrade due to the modification.
We sample two sets of 60K scenarios from the dataset; for each scene, we introduce approximately 25 failures for the first set, and 1 failure per scene for the second set, resulting in a total of \textbf{1.5M} monitored scenes for the entire training set.
We re-evaluate the candidate plans to obtain the reward on the monitored scene, as mentioned before, and categorize each plan into one of three classes based on the reward range: Safe (reward $>225$), Risky (reward $150-225$), and Safety-Critical (reward $<150$). 
\begin{wrapfigure}{c}{0.45\columnwidth}
          \centering
          \vspace{-1em}
         \includegraphics[width=0.45\columnwidth]{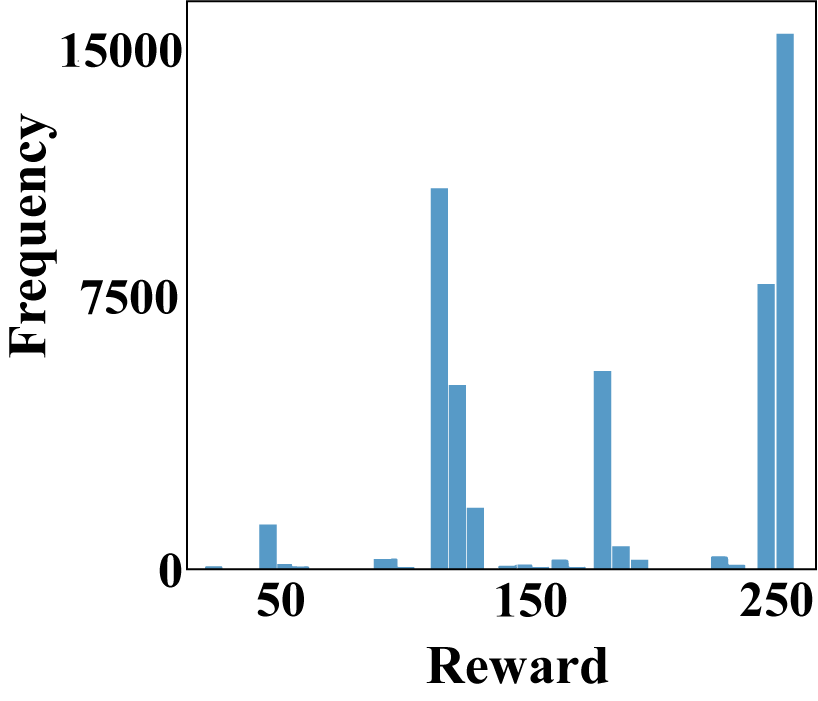}
         \caption{\small{Dataset Reward Distribution.}}
         \vspace{-0.65em}
         \label{fig:reward_dist}
\end{wrapfigure}
These boundaries were determined by analyzing the raw reward distribution in Fig. \ref{fig:reward_dist}.
At runtime, we incorporate the perception monitor's input and utilize \network to estimate the safety classification of the plans under the monitored scene, offering alternative plans as a repair strategy when necessary. 
The network only has access to the original perceived scene, the state of the missing agent as the failure information, and the plan to be evaluated. 

\subsection{Baseline: Hamilton Jacobi Reachability} 
\begin{figure}[ht]
    \centering
    \vspace{0.5em}
         \includegraphics[width=\columnwidth]
         {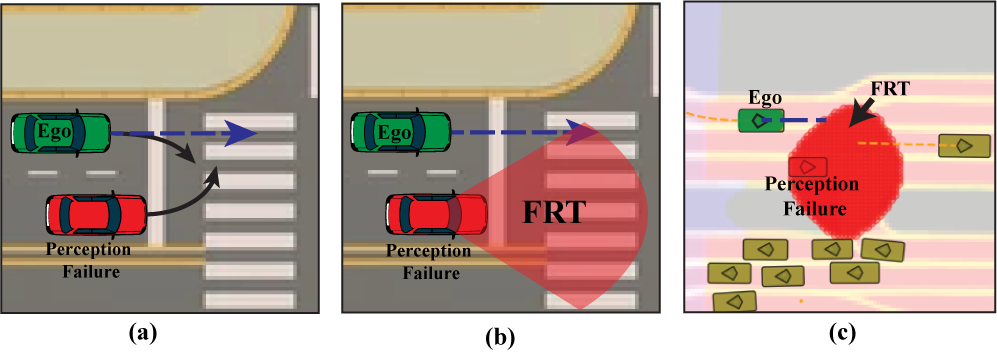}
        \caption{\small{Baselines \textbf{(a)} Two-player game where both the ego and the perception failure choose to maximize the occurrence of collisions. The black trajectories are instantaneous optimal decisions that lead to a possible collision. The blue dashed trajectory depicts the candidate ego plan we wish to analyze for safety. \textbf{(b)} Relaxed baseline where the ego follows the candidate plan, which we wish to check, while we compute the FRT for the perception failure.  
        \textbf{(c)} FRT projected on a random scenario from our dataset.}}
 \label{fig:baselines}
\end{figure}

We compare our approach to different flavours of reachability-based baselines with increasing level of conservativeness. These baselines are chosen to prioritize the safety of the AV and assess the effect of safety over the performance when compared to \network.
\begin{enumerate}
    \item \textbf{Forward Reachable Tube (FRT)} Here, we compute the forward reachable tube (FRT) of the agent that caused the perception failure. The FRT of an agent is the set of all possible states that the agent can reach within a time horizon. If the candidate plan intersects with FRT, the plan is classified as unsafe; otherwise, it is declared safe. We compute the FRT assuming that the missed agent follows unicycle dynamics $f$, which governs the state evolution as follows: 
    \vspace{-0.6em}
    \begin{equation}
    \label{eqn:dyn_dubins}
         \dot \state = f(\state,\ctrl) = \bigr[ v\, cos(\theta)\quad v\, sin(\theta)\quad   \omega \quad a \bigr]
         \vspace{-0.6em}
    \end{equation}
    where $\state$ is the state of the missed agent composed of the xy-position, and heading $\theta$ and $v$, the car's velocity. The control actions are given by $u = \big[\omega\quad a\big]$, where $\omega$ is the steering rate, and $a$ is the acceleration. In addition, we compute a three-dimensional reduction of the model, which assumes a constant velocity ($a = 0$). 
    We use DeepReach \cite{bansal2021deepreach} to approximate a value function whose sub-zero level set gives us the FRT of the agent; additionally, for the 3D system with zero acceleration, we compute a parameterized value function (parameterized by different velocities $v$) whose subzero-level again gives us the FRT. 
    An example FRT is shown in Fig. \ref{fig:baselines}(b)-(c), which flags the ego plan as unsafe since it enters the FRT.
    \item \textbf{Two Player Cooperative Game (2P-Game)}. For our second baseline, we compare our approach to a reachability-based baseline inspired by previous work \cite{topan2022interaction}. In this baseline, the ego and the missed perception failure are involved in a two-player dynamic game where both the players choose to act to maximize the chances of collision using any dynamically feasible trajectory. Here, both the agents follow the unicycle dynamics from eqn. \eqref{eqn:dyn_dubins}. An example of a possible trajectory choice made by the agent and the perception failure in any instance of time is shown Fig. \ref{fig:baselines}. We utilize the framework used in \cite{antonante2023task} to compute the Backward Reachable Tube (BRT), which is the set of all possible initial states from which the system inevitably enters a collation state after a fixed interval of time (3s in our example).  
\end{enumerate}

%
A key distinction between the choice of the baseline is that the FRT-based approach utilizes the candidate ego plan as part of the safety check; in contrast, the 2P-Game version requires that the ego and the agent follow an optimal control strategy that encourages failure. Hence the 2P-Game version cannot be used as a monitor to assess the safety of a candidate plan. On the other hand, due to the closed-loop nature of the game, the 2P-Game version is less conservative as it accounts for the temporal aspects of the interaction, while the instersection of a candidate plan with the FRT does not signify a guaranteed collision. For instance, even if the ego trajectory is flagged as unsafe due to the intersection with the FRT, there is no guarantee that the ego and the agent will reach the exact spatial locations simultaneously (and hence avoid a collision), leading to a more conservative estimate for the safety check. Whereas by definition, states flagged as unsafe by the 2P-Game are certified to lead to a collision within a fixed time interval. 

\begin{table*}[h!] 
\centering
\vspace{0.5em}
\begin{tabular}{l c c c c c}
\hline
\textbf{Algorithm} & \textbf{F1 Score} $\uparrow$ & \textbf{Accuracy} $\uparrow$ & \textbf{Precision} $\uparrow$ & \textbf{Recall} $\uparrow$ & \textbf{Runtime [Hz]} $\uparrow$ \\
\hline
\textbf{SPARQ (ours)}  & \textbf{0.91 (0.81)} &\textbf{ 0.90 }(0.92) & \textbf{0.93 (0.72)} &\textbf{ 0.90} (0.92) &  42  \\
\textbf{Two-Player Game} & 0.57 (0.37) & 0.75 (0.94) & 0.61 (0.23) & 0.75 (0.94) & \textbf{100} \\
\textbf{FRT (3D)} & 0.77 (0.51) & 0.75 (0.92) & 0.85 (0.35) & 0.75 (0.92) & \textbf{100} \\
\textbf{FRT (4D)} & 0.60 (0.46) & 0.80 \textbf{(0.99)} & 0.65 (0.30) & 0.80 \textbf{(0.99)} & \textbf{100} \\
\hline
\vspace{-0.5em}
\end{tabular}
\caption{Performance of \network and reachability-based baseline on the nuPlan-Vegas validation dataset.}
\label{tab:results}
\vspace{-1em}
\end{table*}

\subsection{Quantitative Performance} We assess the performance of \network on the validation 
\begin{figure*}[hb]
    \centering
    \vspace{-0.7em}
         \includegraphics[width=0.83\textwidth]
         {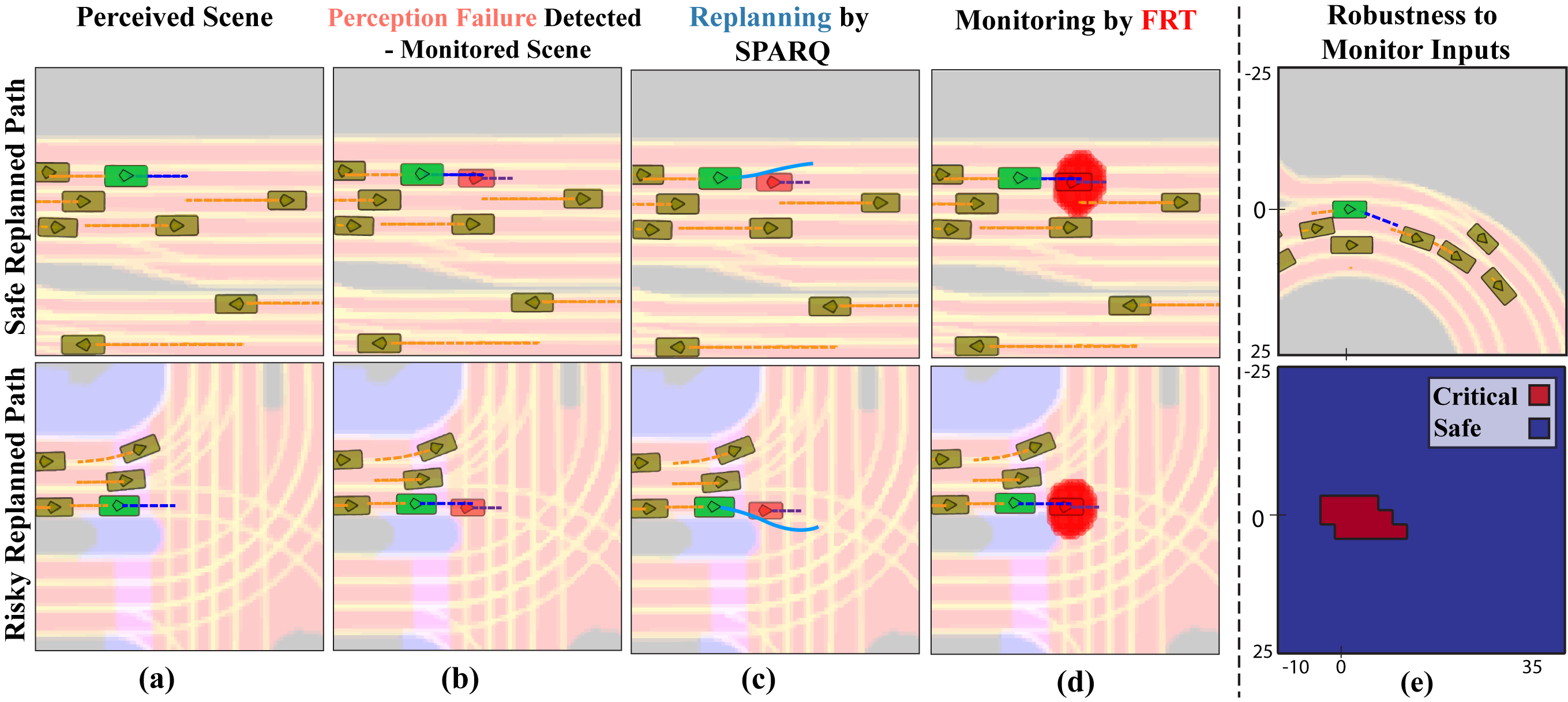}
        \caption{\small{\textbf{(a-d) Plan Repair.} Results showing our algorithm preventing a possible collision by detecting the candidate plan as ``Critical" and subsequently returning a repair plan that was either of the ``Safe" class  \textbf{(top row)} by lane switch maneuver or ``Risky" class \textbf{(bottom row)} via going off-road to perform overtaking action. In both cases the baseline (column (d)) was able to detect the base plan as ``Critical". \\
        \textbf{(e) Robustness to monitor heatmaps}. (\textbf{top}) Shows the perceived scene and (\textbf{bottom}) shows the \network predictions for each location of the monitor input. Red corresponds to that particular location being classified as \textit{Critical}, while Blue is classified as \textit{Safe}.}}
 \label{fig:qualitative}
\end{figure*}
split of the nuPlan-Vegas dataset. The evaluation uses four different metrics: accuracy, precision, recall, and F1 score. Accuracy measures the correctness of the detections, precision measures the quality of positive detections (soundness), recall measures the ability to capture all positive detections (completeness), and F1 score balances both precision and recall to provide a single performance measure for detection.
The metrics are averaged over the three prediction classes.
Since this work prioritizes AV safety, we also compare the methods over their performance on the safety-critical class; the values corresponding to this metric are mentioned in parentheses along with the average metrics. 

The results are presented in Table \ref{tab:results}. Overall, we see that \network either equals or does better than the baselines for almost all the metrics. The FRT and the Two-Player Game baseline exhibit a high recall for the safety-critical class, which is essential for ensuring the safety of autonomous systems. This is not surprising as the Reachability-based methods are known to be conservative. However, it comes at a cost of loss in performance, as evident by the lower precision. This results from a high false positive rate for the safety-critical class for this method. 
Hence, the plans that are actually safe, despite the presence of perception errors, are classified as unsafe. This worst-case assumption approach further leads to the rejection of all the proposed plans, ultimately halting the vehicle as an aggressive option to enforce safety. 
On the other hand, \network balances safety and performance as shown by the consistent performance over all the metrics. Furthermore, we computed a strong AUROC score of \textbf{0.9} by \network for all three classes, nearing the performance of an ideal detector (AUROC = 1). 
In our computation, we pre-calculated the value function for the reachability-based methods, resulting in a lookup table that could be efficiently queried for any state at runtime.

Interestingly, \network demonstrated an unexpected ability to detect failure cases not caused by perception failures, functioning as a general-purpose safety filter. It filtered \textbf{96\%} of plans from the base planner that would have led to safety violations, even without perception failures. We hypothesize that the diversity of our dataset enabled \network to identify fundamental features causing safety failures. 

\begin{remark}
    It was interesting to observe that in spite of being a conservative approach, the reachability-based methods were unable to detect some critical failure scenarios. Additionally, the two-player game version has a lower recall than the FRT-based baseline. This was a direct consequence of using the non-holonomic unicycle dynamics for the prediction model for the FRT and the two-player-based baselines, while the ground truth was computed using differentially-flat dynamics involving motion primitives. The reachable tube for a system obeying a simpler dynamics model(e.g., the motion primitives) is generally larger when compared with the same system with additional non-holonomic constraints (e.g., the unicycle model). The two-player game baseline uses the unicycle dynamics for both the ego as well as the missed agent, while the FRT-based baseline uses it once for the missed agent. SPARQ  is able to leverage the flat dynamics model from the trajectory tree planner to its advantage, resulting in a more accurate safety monitor.
\end{remark}

\subsection{Qualitative Performance} 
We now analyze two cases of plan repair exhibited by our algorithm in Fig. \ref{fig:qualitative}(a-d). In both cases (presented in the top and bottom rows, respectively), our algorithm was able to correctly detect that the candidate ego plan (obtained by \rhplanner~on the perceived scene in column (a)) would be compromised due to the presence of the unaccounted perception failure exposed by the monitor (shown with the red marker in column (b) on the monitored scene). In the \textbf{first row} \network was able to propose a repaired plan that is in the ``Safe" class and was able to prevent crashing into the missed agent by executing a lane change to the left. A lane switch to the right would have again resulted in a critical situation due to the presence of another agent on that side, and indeed, such a plan was not predicted as safe by SPARQ. In the \textbf{second row}, however, \network was unable to propose a plan from the safe class given that the ego vehicle faced a more constrained environment with a curb on its right and another agent to its left, making this a \emph{hard} scenario. Thus, \network proposes a repair from the ``Risky" class. This plan leads the ego vehicle off-road and \textit{overtakes} the missed agent. 
While this plan violated the rule of keeping the AV on the road, it still averted a catastrophic failure -- a collision with another agent. In both cases, being a conservative approach, the baseline method predicted that the candidate plan would enter the FRT of the missed agent and, hence, flagged it as unsafe. However, the maneuvers suggested by \network are also deemed unsafe under FRT, with coming to a complete stop being the only safe plan.   

Finally, we show the robustness of our algorithm with respect to different inputs of the safety monitor. Essentially, we want to visualize all possible locations of a missed agent that are critical to the ego plan. Fig. \ref{fig:qualitative} (column e)  shows a perceived scene (Fig. \ref{fig:qualitative}e \textbf{top}) and the corresponding predictions of \network for each possible location of the missed agent in the scene (Fig. \ref{fig:qualitative}e \textbf{bottom}). The network predicts that the presence of the perception failure around the ego position will result in its plans being classified as ``Critical" (shown in red). In contrast, missing agents at the periphery of the map do not affect the ego plan, and hence, the plan would still be safe, as predicted by \network (``Safe" class predictions are shown in blue).

\noindent \textbf{Runtime details.} One of the key advantages of \network is its significantly smaller inference time, making it feasible for real-time deployment in AV stacks. We demonstrated that our method can classify the task-relevant safety score of approximately 256 plans in 0.024s (or $\sim$\textbf{42Hz}) for a given scenario.
Additionally, it requires only 0.006s to propose a safer alternative plan if the original candidate plan is deemed unsafe. In contrast, the default \rhplanner-based pipeline takes around 0.05s (or \textbf{20Hz}) to perform the same task. This speedup is primarily due to replacing analytical components with neural network-based elements.

Another strength of our method is its modularity. Once trained, \network can be seamlessly integrated with any other planner, serving as a filter for unsafe plans and enabling plan repair under perception failure.

\section{Conclusion and Future Work}
\label{sec:conclusion}
In this paper, we present SPARQ, a Q-network-based approach for run-time monitoring of system-level 
perception failures \emph{and} generating recovery plans. Furthermore, SPARQ doubles down as a general-purpose safety monitor in the absence of any perception failures. In our experiments, SPARQ delivered a low false-positive and false-negative rate while being run-time capable for modern AV stacks.

As a part of our future work, we will explore injecting out-of-distribution (OOD) robustness in the monitor which would allow it to serve as a run-time monitor despite distribution shifts, such as geographical locations, driving behaviors of other agents, etc. Another exciting opportunity is the exploration of natural-language-based descriptions as inputs to the encoder to facilitate multi-modal reasoning in SPARQ.



\bibliographystyle{IEEEtran}
\bibliography{references}

\end{document}